\documentclass[runningheads]{llncs}

\usepackage{graphicx}
\usepackage{subfigure}

\usepackage{diagbox}
\usepackage{threeparttable}
\usepackage{multirow}

\usepackage[ruled]{algorithm2e}

\usepackage{amsmath}

\newcommand{\unprotectedS}{\mathit{u}} 
\newcommand{\protectedS}{\mathit{p}} 

\begin{document}

\title{FARF: A Fair and Adaptive Random Forests Classifier}
\titlerunning{FARF: A Fair and Adaptive Random Forests Classifier}
%
\author{Wenbin Zhang\inst{1} \and
Albert Bifet\inst{2,3} \and
Xiangliang	Zhang\inst{4} \and \\
Jeremy C. Weiss\inst{5} \and
Wolfgang Nejdl\inst{6}
}


\authorrunning{W. Zhang, A. Bifet, X. Zhang, J. C. Weiss and W. Nejdl}
%
\institute{University of Maryland, Baltimore County, MD 21250, USA \and
University of Waikato, Hamilton 3216, New Zealand \and T\'el\'ecom Paris, Institut Polytechnique de Paris, Palaiseau 91764, France \and
King Abdullah University of Science and Technology, Thuwal 23955, Saudi Arabia \and
Carnegie Mellon University, Pittsburgh, PA 15213, USA  \and
L3S Research Center \& Leibniz University Hannover, Hannover 30167, Germany\\
\email{$^1$wenbinzhang@umbc.edu, $^{2,3}$abifet@waikato.ac.nz, $^4$xiangliang.zhang@kaust.edu.sa, $^5$jeremyweiss@cmu.edu, $^6$nejdl@L3S.de}}
\maketitle              

\begin{abstract}
	As Artificial Intelligence (AI) is used in more applications, the need to consider and mitigate biases from the learned models has followed. Most works in developing fair learning algorithms focus on the offline setting. However, in many real-world applications data comes in an online fashion and needs to be processed on the fly. Moreover, in practical application, there is a trade-off between accuracy and fairness that needs to be accounted for, but current methods often have multiple hyper-parameters with non-trivial interaction to achieve fairness. In this paper, we propose a flexible ensemble algorithm for fair decision-making in the more challenging context of evolving online settings. This algorithm, called FARF (Fair and Adaptive Random Forests), is based on using online component classifiers and updating them according to the current distribution, that also accounts for fairness and a single hyper-parameters that alters fairness-accuracy balance. Experiments on real-world discriminated data streams demonstrate the utility of FARF. 
\end{abstract}

\section{Introduction}
AI-based decision-making systems are routinely being used across a wide plethora of online (e.g., the targeting of products, the setting of insurance rates) as well as offline services (e.g., the issuing of mortgage approval, the allocation of health resource). As AI becomes integrated into more systems, various AI-based discriminatory incidents have also been observed and reported~\cite{beutel2019putting,hardt2016equality,meyer2018amazon}. 

A large number of methods have been proposed to address this issue, ranging from discrimination discovery to discrimination elimination and interpretation in order to provide ethical and accurate decisions~\cite{zhang2019fairness,zliobaite2015survey}. These studies have typically adopted one or more of the three following strategies: i) \emph{Pre-processing solutions} aim to eliminate discrimination at the data level, including the most popular ones massaging~\cite{kamiran2009classifying} and reweighting~\cite{calders2009building}. ii) \emph{In-processing approaches} mitigate bias by modifying the algorithm design~\cite{bhaskaruni2019improving,kamiran2010discrimination}. As a recent example, the Bayesian probabilistic modeling is leveraged to account for fairness~\cite{foulds2020bayesian}. iii) \emph{Post-processing techniques} consist of a-posteriori adjusting the output of the model~\cite{hardt2016equality,iosifidis2019fae}. For instance, the decision boundary for the protected group is shifted based on the theory of margins for boosting~\cite{fish2016confidence}.  

However, most of these methods tackle fairness as a static problem, i.e., that all the data is available at training time. This does not satisfy situations that may require online learning due to a continuously drifting data distribution, or can not computationally afford to process all of their data in memory~\cite{zhang2017hybrid}. There is very little work in the area of online learning that includes any definition of fairness as a goal of the method~\cite{iosifidis2019fairness,zhang2019faht}. Our work seeks to fill this void.

Current methods also lack a mechanism for easily adjusting the trade-off that exists between accuracy and fairness \cite{Kleinberg2016}. For instance, the ``business necessity'' clause~\cite{barocas2016big} states that a certain degree of disparate impact discrimination can be allowed for the sake of meeting certain performance-related business constrains, on the condition that such decision-making causes the least disparate impact when fulfilling the current business needs. If an initial model fails to meet the discrimination or accuracy requirement for practical use, we would prefer there exist a single parameter with a direct and predictable impact on this trade-off. However, current studies solely focus on preserving prediction performance while minimizing discrimination, and do not allow for fine-grained control between fairness and accuracy~\cite{beutel2019putting,zliobaite2015survey}.

To overcome these issues we propose FARF, an online statistical parity aware Random Forest (RF) model. Like prior online RF algorithms, it is built from a sampling approach for the ensemble creation. In creating this fair variant of RF, we develop a number of contributions: i) We study a new research direction of fairness-aware learning considering concept and fairness drift. We then propose FARF, a fairness-aware and fairness-updated ensemble method to tackle online fairness. ii) We study another research direction of fairness-aware learning with customized control, and design a clear mechanism for fine-grained fairness control, providing more flexibility than state of the  art. iii) We theoretically analyze the inadequacy of current sampling approaches in fairness studies and introduce a new effective sampling direction with experimental verification. iv) Extensive experimental evaluation on real-world datasets demonstrates the capability of the proposed model in online settings.

\section{Problem Definition}
\label{sec: definition}

An online stream $D$ consists of a sequence of instances arriving over time, potentially infinite. One instance $x_t$ at time step $t$ in $D$ is described in a feature space $A=\{A_1, ..., A_n\}$ within respective domains $dom(A_i)$ and its class label $C_t$. An online classifier is trained incrementally by taking  instances up to time $t$ to predict  $C_{t+1}$ for the unlabeled instance arriving at time step $t+1$. Once $C_{t+1}$ is predicted, the actual class label of $x_{t+1}$ becomes available and can be used for model update, known as prequential evaluation~\cite{gama2010knowledge}.

We assume one of the attributes $A$ is a special attribute $S$, referred to as $sensitive~attribute$ (e.g., gender) with a special value $s \in dom(S)$ referred to as $sensitive~value$ (e.g., female), from which the discriminated group is defined. For simplicity, we consider binary classification tasks assuming $dom(C) \in \{+, -\}$ and $S$ also is binary with $dom(S) \in \{\protectedS, \unprotectedS\}$ (i.e., protected and unprotected respectively). Four fairness related groups can therefore be distinguished combining $S$ and $C$. These groups are $\protectedS^+, \protectedS^-$ and $\unprotectedS^+,  \unprotectedS^-$ representing protected group (e.g., female) receiving positive and negative classification and unprotected group (e.g., male) receiving positive and negative classification, respectively.

Although more than twenty notions have been proposed to measure the discriminative behavior of AI models~\cite{verma2018fairness}, formalizing fairness is a hard topic per se, and there is no consensus which measure is more versatile than others~\cite{beutel2019putting}. In addition, what constitutes ``fair'' or ``discriminative'' is dependent on many factors and context, as well as philosophical questions that have been researched long before the AI communities’ interest~\cite{binns2018fairness}. In this work, we adopt the \emph{statistical parity} because American user studies have found that it is a measure compatible with
many users' intuition of what constitutes a ``fair'' decision~\cite{srivastava2019mathematical}, expecting a wide spectrum of applications of our method. Briefly, statistical parity examines whether the probability of being granted for a positive benefit (e.g., the provision of health care) is the same for both protected and unprotected groups. While statistical parity is designed for offline fairness, the discriminative behavior of the AI model up to time $t$ in the online setting, which we term as~\emph{accumulated statistical parity}, can be analogically defined as:

\begin{equation}
	\label{equ: discrimination}
	Disc(D_t)= \frac{\unprotectedS_{t}^{+}}{\unprotectedS_{t}^{+} + \unprotectedS_{t}^{-}} - \frac{\protectedS_{t}^{+}}{\protectedS_{t}^{+} + \protectedS_{t}^{-}}
\end{equation}

\noindent where $\unprotectedS_{t}^{+}$, $\unprotectedS_{t}^{-}$, $\protectedS_{t}^{+}$ and $\protectedS_{t}^{+}$ are up to time $t$ the number of individuals from respective groups.

People from the protected group can claim they are discriminated up to time $t$ when more of them are rejected a benefit comparing to the people of the unprotected group. The aim of online fairness-aware learning is therefore to provide real time accurate but also fair predictions from the massive data streams, where $D$ needs to be processed on the fly without the need for storage and reprocessing, and data distribution including $Disc(D_t)$ could also evolve over time.

\section{The Fair and Adaptive Random Forests}
\label{sec: method}

Ensemble learning combines multiple base learners to generate more robust descriptions. Three common strategies are bagging, boosting and random forests. Specific to online learning, there are multiple versions of bagging and boosting that are part of the state of the art ensemble methods for evolving online learning~\cite{bifet2010leveraging,chen2012online}, while random forests for non-stationary data stream are currently represented by~\cite{abdulsalam2008classifying,gomes2017adaptive}, which also show random forests approaches have a superior performance comparing to bagging and boosting methods. One possible reason is that training on sampled data and selected features for splitting generalize more than adding more random weights to instances by bagging and adding weights to incorrectly classified instances by boosting. In this paper, we follow the idea of online random forests~\cite{abdulsalam2008classifying,gomes2017adaptive} as a powerful tool to increase the generalization and fairness when constructing an ensemble of classifiers.

Specifically, the proposed Fair and Adaptive Random Forests (FARF) is an adaptation of the classical random forest algorithm~\cite{breiman2001random}, and can also be viewed as an updated and fairness-aware version of the previous attempts to perform this adaptation~\cite{abdulsalam2008classifying,gomes2017adaptive}. In comparison to these attempts, FARF proposes a theoretically sound and fairness-oriented sampling (Section~\ref{sec:sampling}), an updated adaptive strategy~(Section~\ref{sec:alg}) as well as employing a fairness-aware base learner also for ensemble diversity~(Section~\ref{sec: baseLearner}) to cope with discriminatory evolving data streams collectively. The following subsections elaborate these three improvements one by one.

\subsection{Diversified Fairness-aware Base Learner}
\label{sec: baseLearner}

Most of the existing online ensemble approaches~\cite{abdulsalam2008classifying,gomes2017adaptive} induce their base learners based on the Hoeffding Tree (HT) algorithm~\cite{domingos2000mining}, which exploits the fact that an optimal splitting attribute can be determined by a small sample and the learned model is asymptotically nearly identical to that of a conventional non-incremental learner. However, such induction is based on the \emph{information gain (IG)} aiming to optimize for predictive performance and does not account for fairness. In our previous work~\cite{zhang2019faht}, the \emph{fair information gain (FIG)} is proposed as an alternative tree splitting criterion to address the discrimination issue of $IG$, formally put,

\begin{equation}
	\label{equ: fig}
	FIG(D,A) = \left\{
	\begin{array}{lr}
		IG(D,A),  ~~~~~~~~~~\textrm{if}~ FG(D,A)=0 \\
		IG(D,A) \times FG(D,A), ~~~ \textrm{otherwise}
	\end{array}
	\right.
\end{equation}
\noindent where \emph{fairness~gain} ($FG$) measures the discrimination difference due to the splitting and is formulated as:

\begin{equation}
	\label{equ: FEG}
	FG(D,A)= |Disc(D)|-\sum_{v \in dom(A)}|Disc(D_v)|
\end{equation}

\noindent where $D$ is the collection of instances and $A$ represents the attribute that under evaluation, $D_v, v \in dom(A)$ are the partitions induced by $A$, and the resultant discrimination value is assessed according to Equation~(\ref{equ: discrimination}). In $FIG$, multiplication is favoured, when combining $IG$ and $FG$ as a conjunctive objective, over other operations for example addition as the values of these two metrics could be in different scales, and in order to promote fair splitting which results in a reduction in the discrimination after split, i.e., $FG$ is a positive value.

In FARF, other than the discrimination reduction merit similar to the previous fairness-driven $IG$ reformulation efforts~\cite{kamiran2010discrimination,zhang2019faht}, such splitting criterion also detects local discrimination to increase diversity for the sake of maximizing the accumulated fairness. Specifically, each partition induced by the attribute $A$ contributes equally to the accumulated fairness of $A$ regardless the number and size of branches. In the context of ensemble learning, diversity of the each individual classifier plays a key role. Increasing diversity by eyeing on local discrimination, i.e., identifying certain attribute values with a high discrimination rate but small in representation size, could therefore induce diversified base classifiers, reflecting different discrimination representation and improving the final ensemble capability. Such emphasis can also be regarded as selecting those attributes that otherwise would not be used for splitting thus adding more randomization for the construction of the tree.

This diversified fairness-aware learner therefore learns different attribute value level discrimination during the tree construction to maximize the accumulated fairness, and is used as the base learner of FARF. To align with such diversity-promoting strategy, different from the base learner of the previous ensemble approaches~\cite{abdulsalam2008classifying,gomes2017adaptive}, FARF also does not perform early tree pruning for its base learners, and a random subset of fair features are selected for new split attempts to further encourage diversity.

\subsection{Fairness-aware Sampling} 
\label{sec:sampling}

In batch random forests, each base classifier is trained on a bootstrap of the entire training set. However, such bootstrap replicates sampling strategy is infeasible in online setting as each training instance needs to be processed once ``on arrival'' without reprocessing. Oza et al.~\cite{bifet2010leveraging} simulate the construction of bootstrap replicates in online context by sending $K$ copies of each training instance to update the base classifier accordingly, where $K$ is a suitable Poisson random variable. Considering the arbitrary length of online stream, we follow~\cite{bifet2010leveraging} that found setting

\begin{equation}
	\label{equ:k=6}
	\mbox{K} = Poisson(6)
\end{equation}
\noindent to have the best accuracy by increasing diversity of the base learners. Others have consistently found this approach effective in accuracy and computing requirements~\cite{gomes2017adaptive}. Then the latest arriving instances can be classified by voting of the base learners, the same way in online and batch random forests. We will propose two different methods of altering the sampling of $K$ to encourage fair tree induction.

Sampling techniques have been studied in recent fairness-aware learning approaches to alleviate discrimination~\cite{bhaskaruni2019improving,iosifidis2019fae}. In these studies, they exclusively concentrate on \textbf{over-sampling the protected positive group} through different heuristics. However, we argue that such interventions are insufficient especially in online setting for two reasons. First, the protected positive group is normally the under-represented minority. Solely focusing on sparse representation might not have significant bias mitigation effect. Such ineffectiveness is further exacerbated in online setting as instances from the protected positive group could discontinue for a certain period of time. Second, over-sampling protected positive group in random forest can be regarded as \textbf{minority over-sampling with replacement}. Previous research has noted that it does not significantly improve minority class recognition~\cite{chawla2002smote}. We interpret the underlying effect in terms of spreading the decision regions of protected positive group to mitigate biases. Essentially, as protected positive group is over-sampled by increasing amounts, the effect is to learn qualitatively similar but more specific regions that overfit the protected positive group rather than spreading its decision boundary into the unprotected positive group region.

Therefore, instead of over-sampling protected positive group, our ensemble learning method \textbf{under-samples the unprotected positive group} to mitigate the discrimination. We design the update rule for instance weight for sampling as:

\begin{equation}
	\label{equ:fairK}
	fairK(x_t) = \left\{
	\begin{array}{lr}
		Disc(D_t)*K,  \mbox{if $x_t \in$ $\unprotectedS^+$\&$Disc(D_t) > 0$} \\
		K,  ~~~~~~~~~~~~~~~\textrm{otherwise}
	\end{array}
	\right.
\end{equation}

\noindent where $Disc(D_t)$ measures the accumulated discrimination up to the current instance at time $t$ in the stream and $K$ is the Poisson weight defined in Equation~(\ref{equ:k=6}). When the current accumulated discrimination is positive ($Disc(D_t) > 0$), i.e., protected group has been discriminated, and the current instance is a member of unprotected positive group, the sampling weight $fairK (x_t)$ is down-scaled for the current instance $x_t$, making it to be $Disc(D_t)$ proportional of Poisson weight $K$. When there is no membership discrimination against the protected group or  the current instance belongs to unprotected group, $fairK(x_t)$ is equivalent to the Poisson weight $K$. This allows our models to learn a more effective decision surface for the unprotected group, while avoiding prior shortcomings to sampling based fairness.

Other than exclusively focusing on over-sampling the protected positive group, the previous fair sampling studies also require additional neighborhood information through KNN~\cite{bhaskaruni2019improving} and clustering~\cite{iosifidis2019fae}. On the contrary, sampling in our work is directly defined in terms of the targeting discrimination. While enjoying simplicity, this also opens the door to flexible control on the degree of fairness. Specifically, we present a second method of altering the sampling ratio $K$ that allows the user to control a trade off between model accuracy and fairness by manually customizing the re-scaling ratio in $fairK$ to manage the trade-off. This is done with a fixed under-sampling weight $\alpha$ that is incorporated into an alternative equation $customK$ as:

\begin{equation}
	\label{equ:customK}
	customK(x_t) = \left\{
	\begin{array}{lr}
		\alpha*K, & ~~~~~~~~~\mbox{if $x_t \in$ $\unprotectedS^+$ } \\
		K, & \textrm{otherwise}
	\end{array}
	\right.
\end{equation}

\noindent where $\alpha$ is the tunable parameter adjusting the sampling ratio. Note that like $fairK$, the under-sampling only occurs for positive instances of the unprotected group. Such flexible control on the degree of fairness instantiates application-wise fairness-aware learning to accommodate scenarios such as the ``business necessity'' clause~\cite{barocas2016big}.

\subsection{FARF Algorithm} 
\label{sec:alg}

Online fairness additionally requires learning algorithms process each instance upon arrival as well as dealing with non-stationary data distribution indicating concept drifts and fairness implications. That is to say, the relationship between sensitive attribute and class variable might also change over time. A stream classifier pays attention to the boundary evolution but ignores fairness drift. To this end, FARF encapsulates the capability of fairness drift detection and adaptation as well as standby trees and weighted voting to address online fairness comprehensively.

Ensemble learning has been used as a powerful tool by resetting under-performing base learners to adapt to change quickly. The conventional approach resets base learners the moment a drift is detected~\cite{bifet2010leveraging}. However, such reseting could be ineffective since the reseted learner cannot have a positive impact on the ensemble process as it has not been well trained. To this end, FARF employs a more permissive threshold to detect potential drifts and builds standby trees for ensemble members who detect such drifts. The standby trees are trained along the ensemble without intervening the ensemble prediction, and appear on the stage when they outperform their respective ensemble members.

The ensemble design of FARF also offers space for different change detectors being incorporated. One possible detector is ADWIN~\cite{bifet2007learning}, which recomputes online whether two ``large enough'' subwindows of the most recent data exhibit ``distinct enough'' averages, and the older portion of the data is dropped when such distinction is detected. Different from the previous non-stationary studies~\cite{chen2012online,gomes2017adaptive}, FARF employs ADWIN to detect changes in accuracy but also fairness, reflecting both concept and fairness drifts. That is to say drift is detected when either of them evolves.

FARF also weights the prediction of each base learner in proportion to their prequential evaluation~\cite{gama2010knowledge} fairness since its last reset, reflecting the tree performance on the current fairness distribution. Such weighting scheme enjoys the merit of free of predefined window or fading factor to estimate fairness as in other stream ensembles~\cite{abdulsalam2008classifying,gomes2017adaptive} (their estimation focus is accuracy to reflect concept drift though). Note that FARF prioritizes fairness over accuracy by weighting and replacing ensemble members according to fairness. Algorithm~\ref{alg:FARF} shows the sketch of FARF.

\begin{algorithm}[!htb]
	\label{alg:FARF}
	\caption{FARF Leaning Algorithm}
	\LinesNumbered
	\KwIn{a discriminated data stream $D$, the number of base models $M$, optional sampling ratio $\alpha$}
	
	Init base models $h_m$ for all m $\in$ \{1, 2, ..., M\}\\
	\For{each instance $x_t$ in $D$}{
		\eIf{$\alpha$ specified}{
			$w_t \gets $ $customK(x_t)$ according to Equation~(\ref{equ:customK})\;
		}{
			Calculate $Disc(D_t)$ according to Equation~(\ref{equ: discrimination})\;
			$w_t \gets $ $fairK(x_t)$ according to Equation~(\ref{equ:fairK})\;
		}
		
		\For{m= 1, 2, ..., M}{
			Update $h_m$ with $x_t$ with weight $w_t$\;
			\If{ADWIN detects a change in fairness or accuracy in $h_m$}{
				\eIf{standby learner $h'_m = \emptyset$}{
					Build a new diversified fair standby learner $h'_m$\;
				}{
					\If{$|Disc(h_m)|> |Disc(h'_m)|$}{
						Replace $h_m$ with $h'_m$\;
					}
				}
			}
		}
		
		\For{\textbf{all} $h'_m$}{
			Update $h'_m$ with $x_t$ with weight $w_t$\;	
		}	
	}
	\textbf{anytime output:} $h(x_t) = argmax_{c \in C} \sum_{m=1}^{M} W(h_m(x_t) = \mu_m(c))$
\end{algorithm}

For each new instance (line 2), FARF first decides its weight according to fairness-aware sampling based on its fairness information and the accumulated discrimination up to the current instance (line 5-7). When customizable fairness is deployed, the weight is set according to customized sampling ratio (line 3-4). FARF then trains each ensemble member (line 9) with this weight (line 10). When a change is detected (line 11) in one ensemble member who does not have a standby tree (line 12), a respective standby tree is created (line 13), otherwise performances between the ensemble member and its respective standby tree are compared (line 15) to decide ensemble membership replacement if needed (line 16). All standby trees are also trained along the ensemble (line 21-22). The weighted vote can be performed at anytime to predict the class of an instance (line 25). Note that the replacement and voting could also be performed from the accuracy perspective, i.e., replacing the ensemble member when its error is higher and weighted vote on accuracy instead. FARF does fairness replacement and voting in order to prioritize fairness at these steps.

\section{Experimental Evaluation}
\label{sec: experiment}

In the case of static datasets and evaluation, accepted benchmarks for evaluating fairness mitigating approaches are limited in number~\cite{beutel2019putting}. With respect to the highly under-explored online fairness, this challenge is further magnified by the drift and the demanding requirement of the number of instances contained therein. We evaluate our approach on the datasets used in the recent works of this research direction~\cite{iosifidis2019fairness,zhang2019faht}, the $Adult$ and the $Census$ datasets~\cite{Dua:2017} both targeting the learning task of determining whether a person earns more than 50K dollars per annum. We follow the same options in our experiments for fair comparison including the selection of sensitive attribute ``gender'' with female being the sensitive value and processing them in sequence. One difference is that instead of randomizing the order, we order the datasets by the ``race'' attribute for both datasets to better simulate concept drift and possibly increase the learning bias. The previous discussed prequential evaluation is employed for evaluation.

\subsection{Benchmark Performance}
\label{sec:exp_comparison}
\sloppy
This section first investigates the theoretically designed fairness-aware and fairness-updated capabilities of FARF. For comparison, we implemented two recently proposed fair online learners, FEI~\cite{iosifidis2019fairness} and FAHT~\cite{zhang2019faht}. While the paper of FEI did not compare with any baselines, FAHT studied two. We compare with these two baselines therein as well, namely the Hoeffding Tree (HT) and KHT in which the fairness-aware splitting criterion proposed in~\cite{kamiran2010discrimination} is embedded into HT. We also trained the state of the art concept-adapting ensemble learner ARF~\cite{gomes2017adaptive} as another baseline. Other competing fairness methods, including recent proposed fairness ensemble methods which require multiple full data scan, are not considered as none of them can be transferred to online settings. All methods are trained the same way for fair comparison. Relevant results on all datasets are shown in Table~\ref{table: baselines}. Note that since accuracy can be misleading for imbalanced class distributions, we also report Kappa statistics~\cite{gama2010knowledge}.

\begin{table}[!htb]
	\centering
	\caption{The predictive performance-vs-discrimination between FARF and baseline models. Best results in \textbf{bold}, second best in \textit{italics}.}
	\setlength{\tabcolsep}{6pt}
	\label{table: baselines}
		\begin{tabular}{|c|c|c|c|c|c|c|}
			\hline
			\multirow{2}{*}{\diagbox{Methods}{Metric}}& \multicolumn{3}{c|}{\textbf{Adult dataset}} & \multicolumn{3}{c|}{\textbf{Census dataset}} \\ 
			\cline{2-7}
			&   Disc\%  & Acc\% &Kappa\% & Disc\% & Acc\% &Kappa\% \\
			\hline
			HT   &24.14 & 82.16 & 68.15  & 6.61 & 93.11 & 87.54\\
			\hline
			KHT  &  24.24 & 82.43  & 67.2& 6.74 & 93.26 & 87.12\\
			\hline
			FAHT &  \textit{17.20}  & 81.62 &  70.48  & \textit{3.63} & 93.06 & 88.14 \\
			\hline
			ARF   &  24.17 & \textbf{84.51} & \textbf{78.15}  & 6.64 & \textit{94.18} & \textbf{90.41} \\
			\hline
			FEI   &  23.06  & 74.27 & 54.27 & 6.64 & 80.06 & 84.27\\
			\hline
			\textbf{FARF}  & \textbf {8.89}  & \textit {84.19}   & \textit {77.54} & \textbf{0.07}  & \textbf {94.83} &\textit {90.33} \\
			\hline
	\end{tabular}
\end{table}

As shown in Table~\ref{table: baselines} our new FARF method dominates all other baselines in terms of minimizing discrimination, and is best of second-best by both Accuracy and Kappa scores in all other cases. We note that when second best FARF is still highly competitive, being at most 0.78\% within the top performer. This is a desirable trade-off since FARF reduced the discrimination score by a factor of 1.9$\times$ and 51.8$\times$ for Adult and Census dataset, respectively.

\subsection{Accuracy-Fairness Control}
\label{sec:exp_customization}

\begin{figure}[!htb]
	\centering
	\subfigure[Adult dataset]{
		\includegraphics[width=0.45\textwidth]{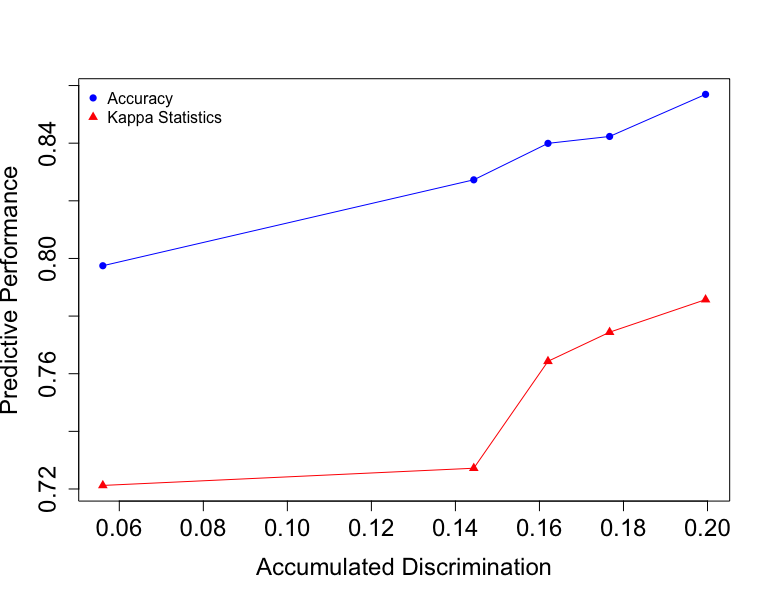}
	}%
	\subfigure[Census dataset]{
		\includegraphics[width=0.45\textwidth]{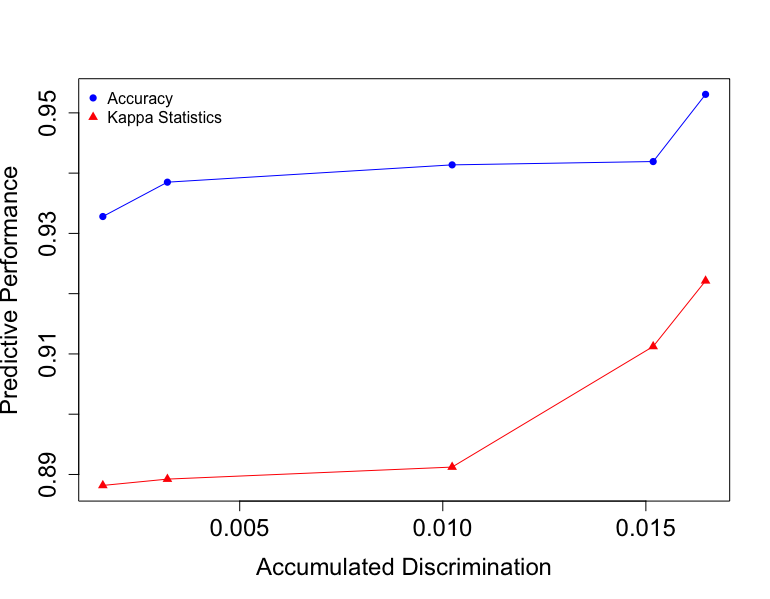}
	}
	\centering
	\caption{The predictive performance and accumulated discrimination trade-off fined-grained by the tunable parameter $\alpha$ ranging from 0.3 to 1.5 with step size 0.3.}
	\label{fig: custom}
\end{figure}

The design of FARF provides a clear mechanism to manage the trade-off between fairness and accuracy. This can be necessary when an initial model does not meet one of these requirements, allowing the end-user to make adjustments. FARF controls thus with the $\alpha$ parameter. As $\alpha$ is in proportion to accuracy, increasing its value leads to a higher accuracy at the expense of a higher discrimination. Such expected trend is clear from the results visualized in Figure~\ref{fig: custom}. Clients can therefore accommodate their needs according to their respective constraints.

The x-axis of the above figure is with respect to the amount of discrimination that is present (larger values indicate more discrimination), and the y-axis is the predictive accuracy (larger is more accurate). With respect to both accuracy and Kappa scores we see a monotonic behavior with respect to the $\alpha$ parameter. This means it behaves as we desire: a simple and direct relationship controlling the trade-off between accuracy and statistical parity. This makes it easy to use, compared to most methods that have multiple parameters that all need to be adjusted to achieve a satisficing trade-off~\cite{Kleinberg2016}.

\subsection{Justification of Sampling Component in FARF}
\label{sec:exp_sampling}

Recent fairness-aware learning approaches employ sampling techniques to mitigate bias, which exclusively focus on over-sampling protected positive group through different heuristics. We theoretically discussed the drawbacks of these methods (c.f., Section~\ref{sec:sampling}). This section provides experimental justification and verifies our choice to instead under sample the protected positive group and that it is critical to our results. We perform two ablations to confirm this by replacing our sampling with: 1)  over-sampling protected positive group, and 2)  over-sampling protected positive group and under-sampling unprotected positive group. All other components of our approach remain the same so that we can isolate our sampling approach as the critical factor in results. These two types of ensemble are denoted as \textbf{FARFS$^-$} and \textbf{FARFS$^{-+}$} respectively in comparison with \textbf{RF}, which refers to random forests without sampling intervention, and our proposed FARF. The results are shown in Table~\ref{table: sampling}.

\begin{table}[!htb]
	\centering
	\caption{The predictive performance-vs-discrimination comparison between different sampling strategies. Best results in \textbf{bold} second best in \textit{italics}. \vspace{-0.2cm}}
	\label{table: sampling}
	\setlength{\tabcolsep}{6pt}
		\begin{tabular}{|c|c|c|c|c|c|c|}
			\hline
			\multirow{2}{*}{\diagbox{Methods}{Metric}}& \multicolumn{3}{c|}{\textbf{Adult dataset}} & \multicolumn{3}{c|}{\textbf{Census dataset}} \\ 
			\cline{2-7}
			&  Disc\%  & Acc\% & Kappa\% & Dis\%  & Acc\% & Kappa\% \\
			\hline
			RF   &  16.32 & \textbf{84.31} & \textbf{78.05}  & 1.34 & \textit{94.13} & \textbf{90.37} \\
			\hline
			FARF$S^-$  &  19.36 & 83.26   & 73.47 & 1.10 & 94.17 & 90.24\\
			\hline
			FARF$S^{-+}$ &  \textit{10.53} & 81.64  &  72.49 &\textit{0.45} & 93.95 & 89.15 \\
			\hline
			\textbf{FARF} & \textbf {8.89}  & \textit {84.19}   & \textit {77.54} & \textbf{0.07}  & \textbf {94.83} &\textit {90.33} \\
			\hline
	\end{tabular} 
\end{table}

As can be seen FARF is the only method that consistently obtains accuracy near that of an unconstrained Random Forest. At the same time, neither approach is able to reach discrimination rates as low as FARF. This shows that over-sampling approaches of prior fairness studies are not as effective as our under-sampling based approach.

\section{Conclusions}
\label{sec: conclusion}

Our work has proposed the first online version of Random Forests with fairness constraints. Our design includes a mechanism for altering the trade off between accuracy and fairness so that users can adjust it easily toward their specific applications. In doing so we have show positive results compared to alternative methods available, without compromising on the desirable properties of online Random Forests.

\bibliography{typeinst}
\bibliographystyle{abbrv}

\end{document}